\documentclass[letterpaper, 10 pt, journal]{IEEEtran}  

\usepackage[ruled,vlined,linesnumbered]{algorithm2e}
\usepackage{amsmath, graphicx } 
\usepackage{balance}

\DeclareMathOperator*{\argmin}{arg\,min}
\usepackage{subcaption} 
\usepackage{amssymb,amsbsy}
\usepackage{breqn,bbm,xcolor}
\usepackage{multirow}
\usepackage[symbol]{footmisc}

\title{\LARGE \bf Pose Estimation for Robot Manipulators via Keypoint Optimization and Sim-to-Real Transfer}

\author{Jingpei Lu$^{1}$ \IEEEmembership{Student Member, IEEE}, Florian Richter$^1$ \IEEEmembership{Student Member, IEEE} \\and Michael C. Yip$^{1}$ \IEEEmembership{Senior Member, IEEE}
\thanks{$^{1}$Jingpei Lu, Florian Richter and Michael C. Yip are with the Department of Electrical and Computer Engineering, University of California San Diego, La Jolla, CA 92093 USA.
{\tt\small\{jil360, frichter, yip\}@ucsd.edu}}%
}

\begin{document}

\maketitle
\thispagestyle{empty}
\pagestyle{empty}

\begin{abstract}
Keypoint detection is an essential building block for many robotic applications like motion capture and pose estimation.
Historically, keypoints are detected using uniquely engineered markers such as checkerboards or fiducials.
More recently, deep learning methods have been explored as they have the ability to detect user-defined keypoints in a marker-less manner.
\textcolor{black}{However, different manually selected keypoints can have uneven performance when it comes to detection and localization.}
An example of this can be found on symmetric robotic tools where DNN detectors cannot solve the correspondence problem correctly.
\textcolor{black}{In this work, we propose a new and autonomous way to define the keypoint locations that overcomes these challenges. The approach involves finding the optimal set of keypoints on robotic manipulators for robust visual detection and localization.}
Using a robotic simulator as a medium, our algorithm utilizes synthetic data for DNN training, and the proposed algorithm is used to optimize the selection of keypoints through an iterative approach. 
The results show that when using the optimized keypoints, the detection performance of the DNNs improved significantly.
We further use the optimized keypoints for real robotic applications by using domain randomization to bridge the reality gap between the simulator and the physical world.
The physical world experiments show how the proposed method can be applied to the wide-breadth of robotic applications that require visual feedback, such as camera-to-robot calibration, robotic tool tracking, and end-effector pose estimation.
As a way to encourage further research in this topic, we establish the ``Robot Pose" dataset, comprising calibration and tracking problems and ground truth data, available online\footnote[2]{Website: https://sites.google.com/ucsd.edu/keypoint-optimization}.
\end{abstract}

\section{Introduction}

\begin{figure}[t]
\vspace{2mm}
    \centering
    \includegraphics[width=0.95\linewidth]{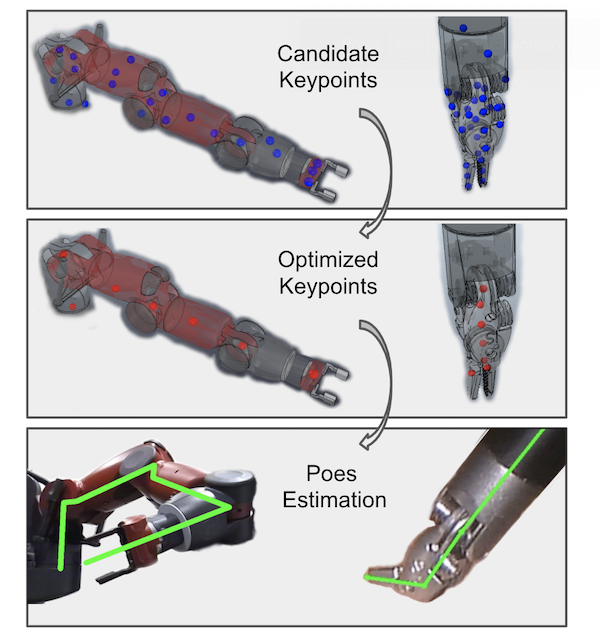}
    \caption{\textcolor{black}{A visualization of the keypoints on the Baxter robot arm (left) and the da Vinci surgical robot tool (right). The candidate keypoints are shown in blue, and the optimized keypoints are shown in red. The bottom row shows these detections being used for pose estimation by utilizing the optimized keypoints.}}
    \label{fig:keypoints_visualization}
    \vspace{-0.1in}
\end{figure}

Visual feedback plays an integral role in robotics because of the rich information images provide.
Historically, there have been two popular approaches to incorporating visual feedback.
The first is through calibration and finding the transform between the base of the robot and the camera \cite{fassi2005hand}.
This transform describes the geometric relationship between objects in the camera frame and the robot, hence allowing image processing and detection algorithms to provide the context necessary for a robot to perform in an environment.
The second approach is directly estimating the relationship between the control, typically joint angles, and the end-effector position in the camera frame \cite{qian2002online}.
This type of visual feedback also allows for end-effector control in the camera frame.

An important step to properly integrating visual feedback techniques is detecting features and finding their correspondence on the robot.
A common approach is placing visual markers on the robot that are easy to detect and hence provide keypoints in the image frame \cite{garrido2014aruco}\cite{olson2011apriltag}.
But where should one place the markers? There does not seem to be any established approach that works the best. One must consider how these marker-based methods require modification of the robot, and how the visibility of the markers will frequently suffer from self-occlusions. Moreover, it is challenging to find the 3D location of the marker relative to the robotic kinematic chain which can cause inaccuracies in visual feedback.
Deep learning approaches for detecting keypoints have been proposed to remove the need for modification of the robot \cite{lu2020super}.
The training of Deep Neural Networks (DNNs) for keypoint detection has even been extended to use synthetically generated data for accurate training with the correct corresponding 3D location relative to the kinematic chain \cite{lee2020dream}.

In spite of the fact that these proposed methods have presented promising ways of keypoint detection, they have failed to address an important consideration which is where to place the keypoints relative to the kinematic chain.
Previous work relies on hand-picked locations for the keypoints which may be sub-optimal and even limiting in performance for the detection algorithm.
An example of this challenge can be found on symmetric robotic tools where the keypoint detection algorithm cannot solve the correspondence problem correctly \cite{lu2020super}.


Taking the advantages of the DNN as the learnable keypoint detector and robotic simulation tools, we present the following contributions:
\textcolor{black}{
\begin{enumerate}
    \item a general keypoint optimization algorithm which solves for the locations of the set of keypoints to maximize their performance on localization tasks,
    \item demonstrations showing that optimized keypoints can improve the performance on real robot pose estimation via sim-to-real transfer, and
    \item methodology for incorporating optimized keypoints with a particle filter to achieving the state-of-the-art performance on surgical tool tracking.
\end{enumerate}
}
We conducted live experiments on both a calibration and a tracking scenario to show the effectiveness of the proposed methods: (i) a Rethink Robotics Baxter robot \cite{robotics2013baxter} for calibrating the robot-to-camera transform and (ii) the da Vinci Research Kit (dVRK) \cite{dvrk} for real-time surgical tool tracking. 
\textcolor{black}{
The datasets for our calibration and tracking experiments are available online. 
The significant kinematic differences between these two robots show the generality of our approach.
We show that the DNN based keypoint detector is capable of consistently and accurately detecting the 2D image projections of optimized keypoints even in cases of self-occlusion. 
The performance using optimized keypoints in both tasks outperforms previous methods with keypoints were selected manually by experienced roboticists, thereby highlighting the usefulness of keypoint optimization.
}

\section{Related Works}

\subsection{Keypoint Optimization}
Early works in robotics have explored how to select the optimal keypoints extracted from SIFT \cite{lowe2004distinctive} and SURF \cite{bay2006surf}, for visual odometry \cite{nannen2013grid} and localization \cite{tamimi2006localization}.
In computer vision, \cite{buoncompagni2015saliency} and \cite{mukherjee2016salient} select the salient keypoints by considering their detectability, distinctiveness, and repeatability.
Recently, there has been a shift to using DNNs for detecting keypoints instead because of the improved performance.
Works in the computer vision community have learned and optimized keypoints for better face recognition and human pose estimation using the DNN \cite{thewlis2017unsupervised}\cite{jakab2018unsupervised}\cite{zhang2018unsupervised}.
However, those algorithms usually consume a large number of real images for training.
Recently, \cite{suwajanakorn2018discovery} proposed an end-to-end framework to optimized the keypoints for object pose estimation in an unsupervised manner by utilizing the synthetically generated data. 
\textcolor{black}{Our work differs from those in the particular goal that optimizing the 3D keypoints for 2D and 3D localization of robot manipulators.} Instead of optimizing the keypoints for specific downstream tasks, our algorithm is more general and can be applied to various robotic tasks that use visual feedback, where the kinematic and the 3D geometric information of keypoints are required. 
\textcolor{black}{
\subsection{Robot Pose Estimation}
A common approach for robot pose estimation is rigidly attaching markers to the robot (e.g. ArUco \cite{garrido2014aruco}), and directly estimate pose from visual data.
The marker-based approach may fail in the case of motion blur and self-occlusion, hence, marker-less approaches are preferable in this scenario.
\cite{schmidt2014dart} and \cite{cifuentes2016probabilistic} utilize depth images for tracking the pose of articulated objects. \cite{lu2020super} and \cite{richter2021robotic} employs DNN to extract point features from RGB images for surgical tool tracking. 
Recent works also address the data labeling issue and investigate using synthetic data to train DNNs for marker-less pose estimation \cite{lambrecht2019towards, lee2020dream}.
None of these previous approaches considered the specific problem of selecting the keypoint locations on robotic manipulators.
We directly compare against previous manually selected keypoint locations for the Baxter robot \cite{lee2020dream} and the daVinci Surgical Robot Research Kit (dVRK) \cite{lu2020super} in our experiments and achieve state-of-the-art performance with our optimized keypoints.
}

\section{Methodology}
\textcolor{black}{
We consider the problem of finding a set of keypoints $\mathcal{P}$ on the robot links that maximize the performance of localizing the robot in both 2D and 3D.
The formulation of this problem is:
\begin{equation}
    \label{eq:argmin_main}
    \mathcal{P}^{\ast} = \argmin_{\mathcal{P}} \mathcal{L}_{2D}(\mathcal{P},\Phi) + \lambda \mathcal{L}_{3D}(\mathcal{P})
\end{equation}
where $\mathcal{L}_{2D}$ is the error of 2D detection with the DNN parameterized by $\Phi$, $\mathcal{L}_{3D}$ is the error of keypoints in the 3D space, and $\lambda$ is a weighting factor.
A set of keypoint is formally defined as $\mathcal{P} := \{\mathbf{p}_{i} | \mathbf{p}_{i} \in \mathbb{R}^3 \}^{K}_{i=1}$, where $\mathbf{p}$ is the 3D position of the keypoint with respect to the robot link it belongs to, and $K$ is the number of keypoints.
To save the efforts of collecting and manually annotating the data, the optimization process is done on the synthetic dataset generated from a robot simulator.
Finally, the keypoint detection in real images is achieved by an effective sim-to-real transfer technique.
}

\subsection{Keypoint Optimization for Robotic Manipulators}
\label{sec:keypoints_optimization}
\textcolor{black}{
Finding the keypoints that maximizing the localization performance will improve many robotic tasks that rely on keypoints as visual feedback.
The DNN is trained end-to-end to detect a set of keypoints, and the detection performance of one certain keypoint can be varied while trained together with other keypoints. This has a significant impact on point-based pose estimation process (e.g. the family of Perspective-n-Point algorithms).
Therefore, the keypoint optimization algorithm's main objective is to find the set of keypoints that optimize the both detection performance of the DNN and the accuracy of 3D pose estimation as shown in (\ref{eq:argmin_main}).
}

\textcolor{black}{
The proposed algorithm assumes $N$ candidate keypoints have existed with known 3D positions along the kinematic chain.
The location of the keypoints can be flexible, as they can randomly distributed to allow non-intuitive keypoint locations.
The size of the keypoint set, $K$, can be varied but should be less than the number of candidate keypoints ($N>K$).
Going through all possible sets of keypoints to solve (\ref{eq:argmin_main}) is intractable.
Instead, we solve for the optimal set by sampling $K$ keypoints, denoted as $\mathcal{P}$, and evaluate their performance on 2D and 3D localization like the loss in (\ref{eq:argmin_main}).
This process is done iteratively where the evaluated performance of each sample set, $\mathcal{P}$, guides future sampling iterations such that the optimal set, $\mathcal{P}^*$, is found without needing to go through all possible sets of keypoints.
The whole optimization process is shown in the Algorithm \ref{alg:keypoints_optimzation}, and each iteration can be broke into four steps: sample keypoints, train model, evaluate performance, and update weights.
}
\begin{algorithm}[t]
\KwIn{$N$ candidate keypoints}
\KwOut{The optimal set of $K$ keypoints $\mathcal{P}^*$}
\tcp*[h]{Initialization} \\
\For {$i = 1$ \KwTo $N$}{
    $w_i^{(0)} = \frac{1}{N}$;
}
$\mathcal{E}_{min} = \infty$;\\
\tcp*[h]{Optimize keypoints iteratively} \\
\textcolor{black}{
\For {$t = 1$ \KwTo $T$} {
     $\mathcal{P}^{(t)} \gets$ \textrm{sampleKeypoints}$(\mathbf{W}^{(t-1)},K)$;\\
     $\Phi^{(t)} \gets$ \textrm{trainModel}$(\mathcal{P}^{(t)})$;\\
     $\mathcal{L}_{total}(\mathcal{P}^{(t)}) \gets$ \textrm{evaluatePerformance}$(\mathcal{P}^{(t)}, \Phi^{(t)})$;\\
     $\mathcal{E}^{(t)} = \frac{1}{K}\sum_{\mathbf{p}_i \in \mathcal{P}^{(t)}} \mathcal{L}_{total}(\mathbf{p}_i)$;\\
     \If{$\mathcal{E}^{(t)} <  \mathcal{E}_{min}$}{
     \tcp*[h]{Update optimal keypoint set} \\
        $\mathcal{E}_{min} = \mathcal{E}^{(t)}$;\\
        $\mathcal{P}^* = \mathcal{P}^{(t)}$;
     }
     $\mathbf{W}^{(t)} \gets$ \textrm{updateWeights}$(\mathcal{L}_{total}(\mathcal{P}^{(t)}), \mathbf{W}^{(t-1)})$;
 }
 }
 \Return{$\mathcal{P}^*$}
 \caption{Keypoint Optimization for Robot Manipulators}
\label{alg:keypoints_optimzation}
\end{algorithm}
\textcolor{black}{
\subsubsection{Sample Keypoints} 
Each candidate keypoint is associated with weight variable $w \in \mathbb{R}$, which can be interpreted as the confidence of being in the optimal set. The $K$ keypoints are sampled based on their weights in each iteration.
The function \textbf{sampleKeypoints}$(\mathbf{W},K)$ takes in the weights of all the candidates $\mathbf{W} \in \mathbb{R}^{N \times 1}$ and randomly selects $K$ keypoints among the candidates according to the probability:
\begin{equation}
\label{eq:sampling}
    P(\mathbf{p}_i) = \frac{w_i}{\sum \limits_{n \in N} w_n}.
\end{equation}
The weights, $w_i$, are iteratively adjusted according to the keypoints 2D and 3D performance hence guiding the search to solve (\ref{eq:argmin_main}).
More advanced sampling can be applied by introducing the constraints. For example, if we want to constraint the sampling so that one keypoint is selected per link, the candidates can be divided into sub-groups for each link.
}
\textcolor{black}{
\subsubsection{Train Model} 
The second step is to train a detection model to detect the sampled keypoint set $\mathcal{P}$.
We utilize the backbone neural network from DeepLabCut \cite{mathis2018deeplabcut} as our detection model and the dataset is synthetically generated from a robot simulator (see Section \ref{sec:domain_randomization} for data generation). We split the dataset into training set and testing set.
The function \textbf{trainModel}$(\mathcal{P})$ trains the DNN on the training dataset to predict the image coordinates of the selected keypoints:
\begin{equation}
    \label{eq:argmin}
    \Phi^{\ast} = \argmin_{\Phi} \mathcal{L}_{train}(\mathcal{P},\Phi) 
\end{equation}
where $\Phi$ denotes the parameters of the detection model and it is optimized using stochastic gradient descent with training loss $\mathcal{L}_{train}$ defined as the cross-entropy loss (see \cite{mathis2018deeplabcut} for details).
In our implementation, we generate the training and testing images with the ground-truth label for all $ N $ candidate keypoints beforehand. By doing so, the labels for selected keypoints in each iteration can be obtained without re-generating the datasets.
}
\textcolor{black}{
\subsubsection{Evaluate Performance} 
The third step is to evaluate the localization performance of the keypoints $\mathcal{P}$ with the detection model $\Phi$ using the testing data on the loss shown in (\ref{eq:argmin_main}). 
We define the 2D detection error, $\mathcal{L}_{2D}$, as the pixel-wise $L_2$ distance (Euclidean distance) between the detection and the ground-truth keypoint:
\begin{equation}
    \label{eq:2D_loss}
    \mathcal{L}_{2D}(\mathbf{p}_i , \Phi) = \frac{1}{M} \sum_{m=1}^{M} || f^{i}(I_m; \Phi) - \mathbf{h}_{i,m}||_2
\end{equation}
where $f^{i}(I_m; \Phi)$ and $\mathbf{h}_{i,m}$ are the DNN detected and ground-truth pixel location respectively of $i$-th keypoint for the testing image, $I_m$, and $M$ is the number of images in the testing dataset.
The 3D keypoint error, $\mathcal{L}_{3D}$, is defined as the average $L_2$ distance between the estimated and ground-truth 3D keypoint position in the camera frame $\{C\}$.
For a keypoint with known position with respect to the $j$-th robot link $\mathbf{p}_i^j$, its position in camera frame can be obtained through forward kinematics and robot-to-camera transformation:
\begin{equation}
    \overline{\mathbf{p}}^{C}_{i} = \mathbf{T}_B^{C} \prod \limits_{n=1}^{j} \mathbf{T}_n^{n-1}(q_n) \overline{\mathbf{p}}^j_i
    \label{eq:forward_kin}
\end{equation}
where $\mathbf{T}_n^{n-1}(q_n)$ is the $n$-th homogeneous joint transform with joint angle $q_n$. Note that coordinate frame 0 is the base frame of the robot and $\overline{\cdot}$ represents the homogeneous representation of a point (e.g. $\overline{\mathbf{p}} = [\mathbf{p}, 1]^T$). The robot-to-camera transformation, $\mathbf{T}^C_B$, is computed through the Efficient Perspective-n-Point Pose Estimation Algorithm (EPnP)\cite{lepetit2009epnp} with the detected keypoints, $f^{i}(I_m; \Phi)$, and their corresponding position in the robot base frame.
Then, the 3D keypoint error is calculated as:
\begin{equation}
    \label{eq:3D_loss}
    \mathcal{L}_{3D}(\mathbf{p}_i) = \frac{1}{M} \sum_{m=1}^{M} || \tilde{\mathbf{p}}_i^C - \mathbf{p}_i^C||_2
\end{equation}
where $\tilde{\mathbf{p}}_i^C$ is the estimated keypoint position from (\ref{eq:forward_kin}) using the estimated $\mathbf{T}^C_B$ from the current keypiont set and $\mathbf{p}_i^C$ is the ground-truth keypoint position in the camera frame.
}

\textcolor{black}{
Using the 2D and 3D losses in (\ref{eq:2D_loss}) and (\ref{eq:3D_loss}) respectively, a total loss similar to (\ref{eq:argmin_main}) per keypoint in $\mathcal{P}$ can be defined.
The function \textbf{evaluatePerformance}$(\mathcal{P}, \Phi)$ does this by simply summing the two losses:
\begin{equation}
    \label{eq:summed_keypoint_loss}
    \mathcal{L}_{total}(\mathbf{p}_i) = \mathcal{L}_{2D}(\mathbf{p}_i , \Phi) + \lambda \mathcal{L}_{3D}(\mathbf{p}_i)
\end{equation}
where $\mathcal{L}_{total}(\mathbf{p}_i)$ is the total loss for $i$-th keypoint.
This loss is considered an evaluation of the keypoint $\mathbf{p}_i$ with respect to the optimization problem in (\ref{eq:argmin_main}) and used to guide future sampling to find the optimal set $\mathcal{P}^*$.
}

\textcolor{black}{
\subsubsection{Update Weights} 
At the end of each iteration, the function \textbf{updateWeights}$(\mathcal{L}_{total}(\mathcal{P}))$ is used to update the weight of the selected keypoints based on (\ref{eq:summed_keypoint_loss}).
At iteration $t$ of the optimization, the weight for the $i$-th keypoint in the sampled set $\mathcal{P}$ is updated to:
\begin{equation}
    w_i^{(t)} = \left(\sum_{k \in K} w_k^{(t-1)} \right) \frac{e^{-\gamma \mathcal{L}_{total}(\mathbf{p}_i)}}{\sum \limits_{k \in K} e^{-\gamma \mathcal{L}_{total}(\mathbf{p}_k)}}
\end{equation}
where $\gamma$ is a tuned parameter to control the effect of the error on changing the weights.
The updated weight will increase or decrease according to the keypoints 2D and 3D performance hence guiding the sampling in \textbf{sampleKeyupoints}$(\mathbf{W}, K)$ towards the optimal set, $\mathcal{P}^*$ that solves (\ref{eq:argmin_main}).
}

\textcolor{black}{
\subsection{Keypoint Detection via Domain Randomization}
\label{sec:domain_randomization}
}

\begin{figure}[t!]
\vspace{2mm}
\begin{subfigure}{0.24\textwidth}
\includegraphics[width=1\textwidth]{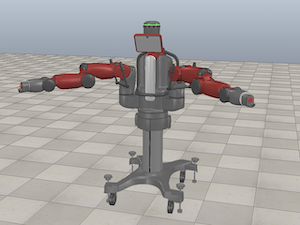}
\vspace{-0.12in}
\end{subfigure}
\begin{subfigure}{0.24\textwidth}
\includegraphics[width=1\textwidth]{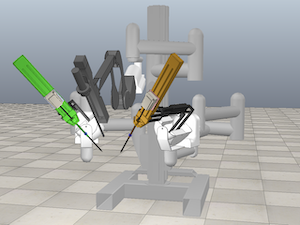}
\vspace{-0.12in}
\end{subfigure}
\\
\begin{subfigure}{0.24\textwidth}
\includegraphics[width=1\textwidth]{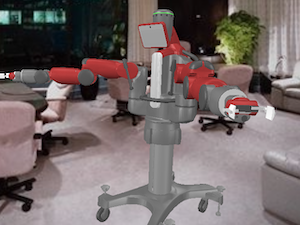}
\vspace{-0.12in}
\end{subfigure}
\begin{subfigure}{0.24\textwidth}
\includegraphics[width=1\textwidth]{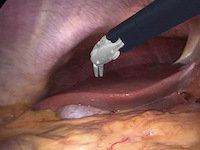}
\vspace{-0.12in}
\end{subfigure}
\caption{Simulation setup and synthetically generated image of the Rethink Baxter (left) and the da Vinci Surgical System (right).}
\label{fig:simulation_setup}
\vspace{-0.1in}
\end{figure}

\textcolor{black}{
A simulated environment is used to generate the synthetic data for the proposed keypoint optimization method.
By using simulated data, ground truth labels, which are crucial to evaluate a keypoints performance in (\ref{eq:2D_loss}) and (\ref{eq:3D_loss}), can be directly generated.}
We set up the simulation environment using the robotic simulator CoppeliaSim \cite{coppeliaSim} and interfaced using PyRep \cite{james2019pyrep} to generate the ground truth label and render RGB images.
Fig. \ref{fig:simulation_setup} shows the simulator view and the rendering view for two robots, where the RGB images are rendered from the virtual cameras. 

\textcolor{black}{
Although the simulator is crucial for ground truth labels and hence ideal for our keypoint optimization method, a naively trained DNN on the labels will be limited to only detecting keypoints on simulated images and may not generalize to the real world.}
To bridge the reality gap, the simple but effective technique known as \textit{domain randomization} \cite{tobin2017domain} is applied to transfer the keypoint detection DNN from the virtual domain to robots in the physical world. 
During the data generation, the virtual cameras are placed in the simulated scene that approximately matches the viewpoint of the real camera, and the following randomization settings are applied to generate the training samples:
\begin{itemize}
  \item The angle for the robot joints is randomized within the joint limits.
  \item The pose of virtual cameras are randomized by adding a zero-mean Gaussian noise to the initial pose, such that
  \begin{equation}
    [\mathbf{q}_{rand}, \mathbf{b}_{rand}]^\top \sim \mathcal{N}([\mathbf{q}_{init}, \mathbf{b}_{init}]^\top, \mathbf{\Sigma} )
    \end{equation}
    where $\mathbf{q} \in \mathbb{S}^3$ is the quaternion, $\mathbf{b} \in \mathbb{R}^3$ is the translational vector, and $\mathbf{\Sigma}$ is the covariance matrix.
  \item The number of the scene lights is randomly chosen between 1 to 3, and are positioned freely in the simulated scene with varying intensities.
  \item Distractor objects, like chairs and tables, are placed in the simulated environment with random poses.
  \item \textcolor{black}{The background of the rendered images are randomly selected from Indoor Scene dataset \cite{quattoni2009indoor} and Hamlyn Centre Endoscopic Video dataset \cite{mountney2010three}.}
  \item The color of the robot mesh is randomized by adding a zero-mean Gaussian noise with a small variance to the default RGB value.
  \item The rendered images are augmented by adding the additive white Gaussian noise using the image augmentation tool \cite{imgaug}.
\end{itemize}
These randomization techniques were applied when generating synthetic data for both keypoint optimization and domain transfer \textcolor{black}{hence ensuring the DNN optimized in (\ref{eq:argmin}) will generalize its keypoint detection to the real world.}



\section{Experiments and Results}
\label{section:experiment}
In this section, we describe efforts towards evaluating the robustness of the keypoints optimization algorithm and the performance of using them in real robotics applications. 
\textcolor{black}{Specifically, we compare our approach with the state-of-art algorithm and marker-based approach on a robot-to-camera pose estimation task and examine the differences between using the optimized keypoints and hand-picked keypoints on a robot tool tracking experiment.
Moreover, we also study the impact of different neural network architectures on the keypoint optimization algorithm.}

\subsection{Datasets and Evaluation Metrics}

\subsubsection{Baxter dataset} 

This dataset contains 100 image frames (resolution: 2048$\times$1526) of the Baxter robot with 20 different joint configurations collected using a Microsoft Azure Kinect. The ground-truth end-effector positions in the camera frame are provided, which is obtained by attaching an Aruco marker physically at the end-effector position. 
The performance of the optimized keypionts is evaluated in both 2D and 3D by estimating the end-effector position.

\textcolor{black}{
For 3D evaluation, we first estimate the robot-to-camera transformation using the optimized keypoints (rotation $\widetilde{\mathbf{R}}^C_B$ and translation $\widetilde{\mathbf{b}}^C_B$). 
Then, we transfer the end-effector from robot base frame to camera frame and calculate the $L_2$ distance between the ground-truth and estimated end-effector position: 
\begin{equation}
    \label{eq:3D_error}
 \| (\widetilde{\mathbf{R}}^C_B\mathbf{x}^B_{ee} + \widetilde{\mathbf{b}}^C_B) - \mathbf{x}^C_{ee}\|_2
\end{equation}
where $\mathbf{x}^B_{ee}$ is the end-effector position in robot base frame obtained by forward kinematics of the robot, and $\mathbf{x}^C_{ee}$ is the ground-truth end-effector position in camera frame measured by Aruco marker.
}

For 2D evaluation, we propose the reprojection error (RE) for the end-effector. The RE is the $L_2$ distance between the estimated end-effector position and its ground-truth position in image coordinates,
\begin{equation}
\textcolor{black}{
    \label{eq:2D_error}
    RE = \left\|\frac{1}{z}\mathbf{K}(\widetilde{\mathbf{R}}^C_B\mathbf{x}^B_{ee} + \widetilde{\mathbf{b}}^C_B) - \overline{\mathbf{h}}_{ee}\right\|_2
}
\end{equation}
\textcolor{black}{
where $\mathbf{h}_{ee}$ is the ground-truth end-effector position in image coordinates, $\mathbf{K} \in \mathbb{R}^{3\times3}$ is the intrinsic matrix, and $z$ is the z-value of the projected point.}
The percentage of correct keypoints (PCK), proposed in \cite{yang2012pck}, is a metric for visualizing the keypoint detection performance across the dataset.
The PCK measures the percentage of keypoints that the distance between the predicted and the ground-truth position is within a certain threshold. 
\textcolor{black}{
In this experiment, the PCK for 2D end-effector localization is calculated in pixels and the PCK for 3D end-effector localization is calculated in millimeters.
}

\subsubsection{SuPer tool tracking dataset}
The SuPer dataset\footnote{https://sites.google.com/ucsd.edu/super-framework/home} is a recording of a repeated tissue manipulation experiment using the da Vinci Research Kit (dVRK) surgical robotic system \cite{dvrk}, where the stereo endoscopic video stream and the encoder readings of the surgical robot are provided. 
We extended the original surgical tool tracking dataset, which originally has 50 ground-truth surgical tool masks, to 80 ground-truth masks by hand labeling. The extended dataset covers a better variation of the tool poses, and the performance of the tool tracking is evaluated using the Intersection-Over-Union (Jaccard Index) for the rendered tool masks,
\begin{equation}
    IoU = \frac{|\mathbf{G} \cap \mathbf{P}|}{|\mathbf{G} \cup \mathbf{P}|} 
\end{equation}
where $\mathbf{G}$ is the ground-truth tool mask area and $\mathbf{P}$ is the predicted tool mask area in the image plane.

\subsection{Keypoints Optimization}

\begin{figure}[t]
\vspace{1mm}
    \centering
    \includegraphics[width=1.0\linewidth]{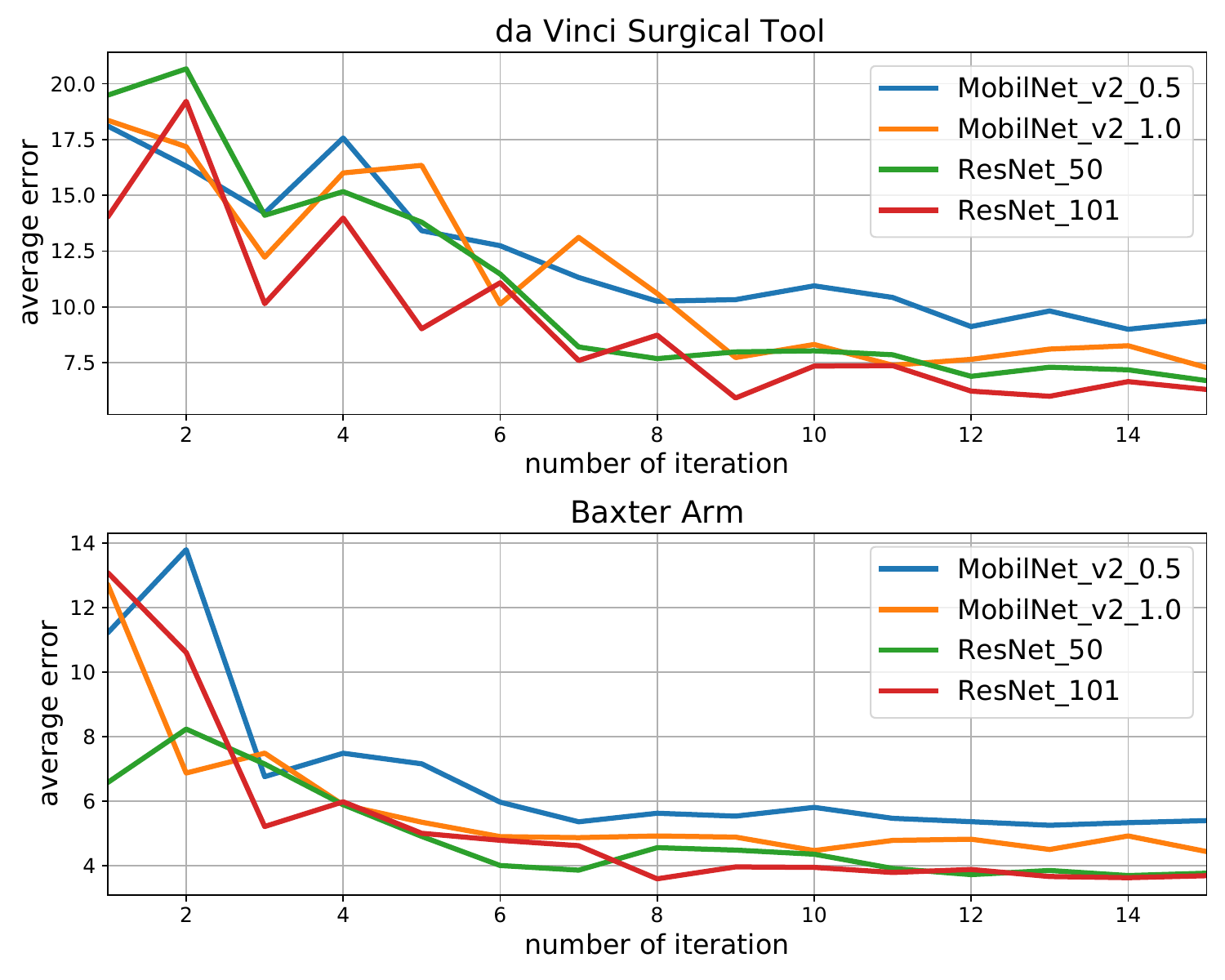}
    \caption{The plot of average error from the \textbf{evaluatePerformance} step for each iteration of keypoint optimization algorithm, \textcolor{black}{while optimizing the keypoints for da Vinci surgical tool (top) and the Baxter arm (bottom).} The algorithm behaves similarly with various DNN architectures.}
    \label{fig:keypont_optimization}
    \vspace{-0.16in}
\end{figure}

\begin{figure}[t]
\vspace{2mm}
    \centering
    \includegraphics[width=1.0\linewidth]{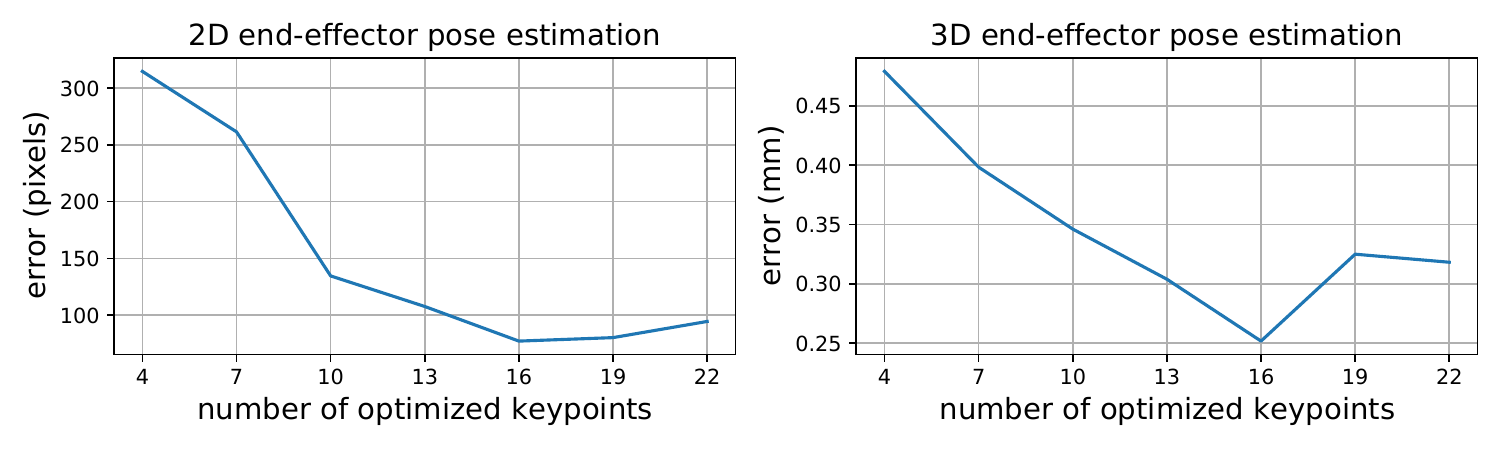}
    \caption{\textcolor{black}{The performance of pose estimation using different numbers of optimized keypoints on Baxter dataset.}}
    \label{fig:err_vs_K}
    \vspace{-0.14in}
\end{figure}

\textcolor{black}{To demonstrate the keypoint optimization algorithm, we randomly placed 32 candidate keypoints ($N = 32$) on the da Vinci surgical tool with a uniform distribution in 3D Euclidean Space and 22 candidate keypoints on the Baxter left arm manually.}
The Algorithm \ref{alg:keypoints_optimzation} is applied to optimize the set of 7 ($K = 7$) keypoints for robust detection. 
The keypoint optimization algorithm was running for 15 iterations ($T = 15$) with $\gamma = 1$ on the synthetic dataset, containing 2K samples for training and 500 samples for evaluations. In the training step, the DNN is trained on the training dataset for 100,000 iterations, with a learning rate of $0.2$. 

To examine the impact of different neural network architectures on the keypoint optimization algorithm, we trained the DNNs with four different feature extractors: ResNet\_50, ResNet\_101, MobileNet\_v2\_1.0, and MobileNet\_v2\_0.5 \cite{mobilenet}. The last digit indicates the number of layers and the width of the network for ResNets and MobilNets respectively, which essentially implies the number of parameters of the network. 
\textcolor{black}{The average keypoint errors from the \textbf{evaluatePerformance} step in each iteration are shown in Fig. \ref{fig:keypont_optimization}, demonstrating that our algorithm is agnostic to different DNN architectures.}
With different feature extractors, the algorithm can converge at a similar rate to the same set of optimal keypoints. 
\textcolor{black}{
The number of iterations required to converge to an optimal keypoint set is around 10 hence showing that our proposed method is substantially faster than brute-force searching through all possible keypoint sets.
}

\textcolor{black}{
We also investigated the choice of $K$, the number of optimized keypoints, for pose estimation. We applied the Algorithm 1 with various $K$ and evaluate the pose estimation performance with the optimized keypoints on the Baxter dataset. We found that the estimation performance gets better in the beginning as the number of optimized keypoints increases. The resulting optimized keypoints are able to create a better coverage of the robot since there are more keypoints. However, when there are too many (i.e. greater than 16 on the Baxter), the low-quality keypoint detections reduce the performance of the overall pose estimation.}

\subsection{Robot-to-camera Pose Estimation from a Single Image}
\label{sec:baxter_experiments}


\begin{figure}[t]
\vspace{2mm}
\centering
\begin{subfigure}{0.22\textwidth}
\includegraphics[width=1\textwidth]{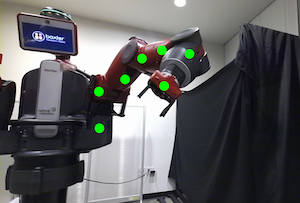}
\vspace{-0.12in}
\end{subfigure}
\begin{subfigure}{0.22\textwidth}
\includegraphics[width=1\textwidth]{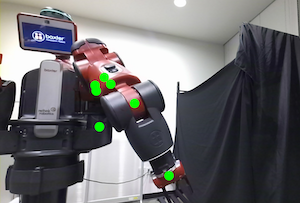}
\vspace{-0.12in}
\end{subfigure}
\\
\begin{subfigure}{0.22\textwidth}
\includegraphics[width=1\textwidth]{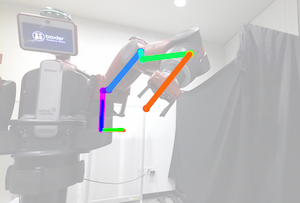}
\vspace{-0.12in}
\end{subfigure}
\begin{subfigure}{0.22\textwidth}
\includegraphics[width=1\textwidth]{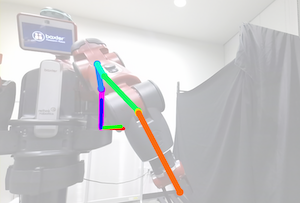}
\vspace{-0.12in}
\end{subfigure}
\caption{\textcolor{black}{The examples of the optimized keypoints detection (top) and the skeletonization (bottom) of the Baxter robot in real-world images. The skeleton of the arms is based on the estimated base frame (RGB reference frame). Keypoints are estimated even with self-occlusions, and the skeletons are perfectly aligned with the robot arm.}}
\label{fig:baxter_real}
\vspace{-0.12in}
\end{figure}

We explored robot-to-camera pose estimation performance from a single RGB image by utilizing the optimized keypoints on Baxter robot.
\textcolor{black}{
The keypoints optimization is constrainted so that one keypoint is sampled for each robot link.
The resulting 7 optimized keypoints are detected on the Baxter dataset to estimate the robot-to-camera transform, $\mathbf{T}^C_B$, on each frame in the same manner as the 3D loss, $\mathcal{L}_{3D}$, from the \textbf{evaluatePerformance}($\mathcal{P}, \Phi $) step of the proposed method.
Note that the ($\widetilde{\mathbf{R}}^C_B$, $\widetilde{\mathbf{b}}^C_B$) used to compute the errors in (\ref{eq:2D_error}) and (\ref{eq:3D_error}) is simply the rotation matrix and translation from $\mathbf{T}^C_B$.
}

We randomly placed three candidate keypoints for each link on the left arm, and the Algorithm \ref{alg:keypoints_optimzation} was applied to find one optimal keypoint per link, \textcolor{black}{with $T = 20, \lambda = 50$, and $\gamma = 2$}.
The ResNet\_50 was utilized as the feature extractor of the detection neural network and was initialized with ImageNet-pretrained weights. The DNNs were trained for 500,000 iterations with a decayed learning rate.
The optimized keypoints for the Baxter's left arm are shown in the middle-left of Fig. \ref{fig:keypoints_visualization}.

To transfer the keypoint detection to the real robot, we applied the domain randomization to bridge the reality gap, as described in Section \ref{sec:domain_randomization}.
The experimental results show that the DNN generalizes well to real-world images.
Fig. \ref{fig:baxter_real} shows the optimized keypoints' detections on the Baxter dataset, demonstrating that the keypoints can be detected even with self-occlusions. Given the accurate keypoint detections, the estimated skeletons are perfectly aligned with the robot arm.

\begin{figure}[t]
\vspace{2mm}
    \centering
    \includegraphics[width=0.95\linewidth]{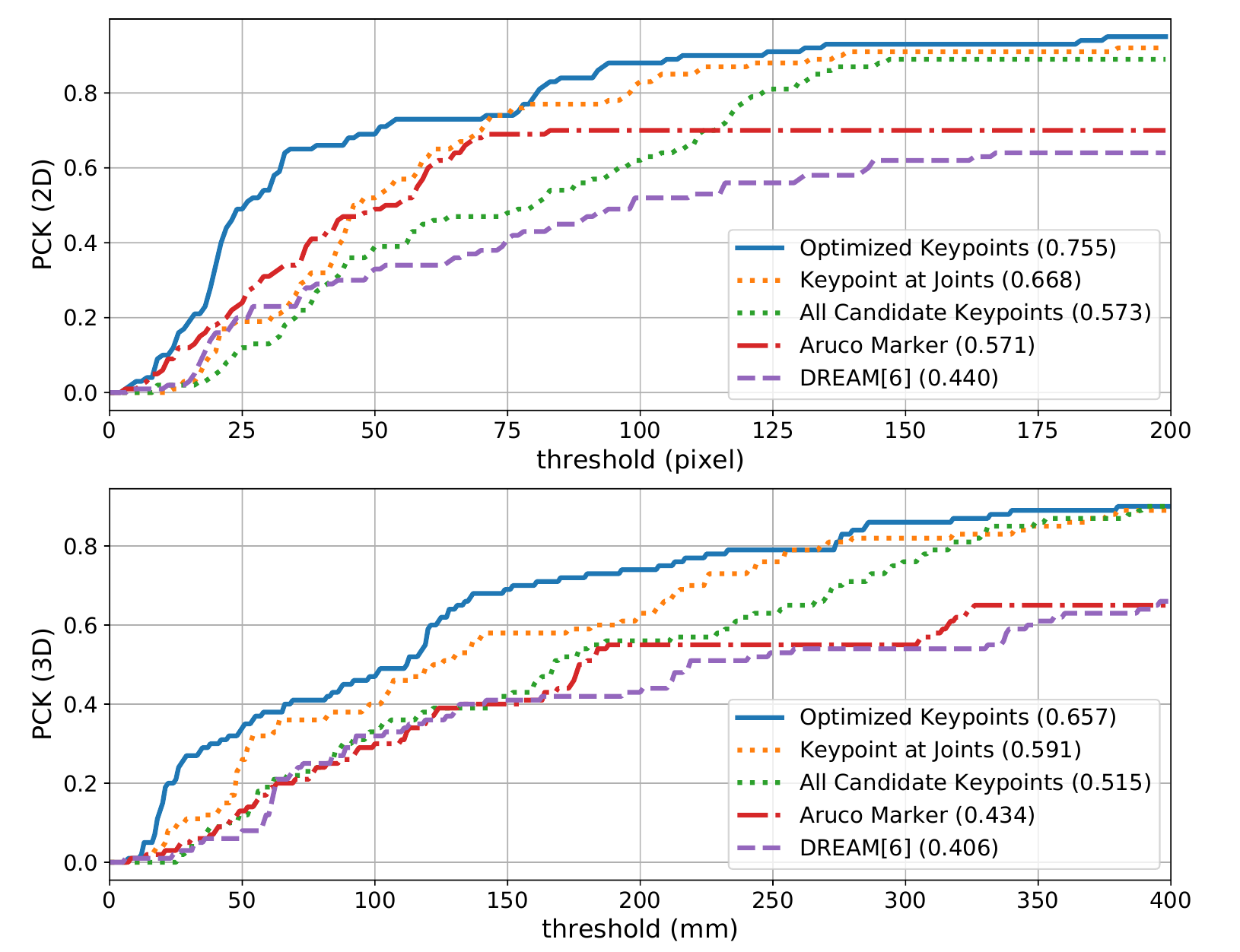}
    \caption{\textcolor{black}{The PCK results on the Baxter dataset for end-effector localization in 2D (top) and 3D (bottom). The thresholds are the $L_2$ distances and the numbers in parentheses indicate the area under the curve (AUC). Around 50 percent of the estimations have an error of less than 1 percent of the image size (2048$\times$1526) using the optimized keypoints.}}
    \label{fig:pck_real}
    \vspace{-0.12in}
\end{figure}

\textcolor{black}{
For comparison, we also experimented the robot-to-camera pose estimation by using all the candidate keypoints or placing the keypoints at the exact joint locations.} 
The state-of-the-art algorithm, DREAM \cite{lee2020dream}, is also implemented for pose estimation which uses robot joints as the keypoints. The 2D keypoints are detected using the DNN from DREAM.
We also compared the traditional camera-to-base pose estimation procedure by placing the Aruco markers on the robot arm. 
The 2D and 3D PCK results for different methods are shown in Fig. \ref{fig:pck_real}.
Using the optimized keypoints, around 50 percent of the estimations have fewer than 25 pixels error in the image plane (about 1 percent of the image size) and have an error of less than 100mm in 3D space, which is much better than other methods.
The area under the curve (AUC), indicating the mean of PCKs, also highlights the overall better performance of using the optimized keypoints.
Due to the self-occlusions and the camera's pose, limiting the visibility of visual markers, some image frames do not have enough detected Aruco markers for EPnP ($< 4$), which hampers the performance.


\subsection{Surgical Tool Tracking}
\label{sec:tool_tracking_experiments}


Surgical tool tracking continuously estimates the 3D pose of the tool end-effector with respect to the camera frame. 
This is necessary for augmented reality displays \cite{richter2019sarpd} and transferring of learning-based control policies \cite{richter2019dvrl}, among other applications \cite{yipDasJournal}. 
We employed the tool tracker previously developed in \cite{lu2020super}, which combines a keypoint detector and a particle filter for 3D pose estimation.
\textcolor{black}{
The keypoint detector (e.g. the proposed optimized keypoint detection method) detects keypoints, $\mathbf{h}_{i, t}$, on the input image at time $t$ to provide visual feedback. 
The corresponding 3D point $\mathbf{p}_i$ is projected to image frame as:
\begin{equation}
    \overline{\mathbf{m}}_{i}(\boldsymbol{\omega}, \mathbf{b}) = \frac{1}{z} \mathbf{K}\mathbf{T}_{B-}^{C} \mathbf{T}^{B-}_B (\boldsymbol{\omega}, \mathbf{b}) \prod \limits_{n=1}^{j} \mathbf{T}_n^{n-1}(q_n) \overline{\mathbf{p}}^j_i
\end{equation}
where $\mathbf{m}_{i}$ is the re-projected keypoint location, $\mathbf{T}_{B-}^{C}$ is the initial hand-eye transform from calibration, and $\mathbf{T}^{B-}_B(\boldsymbol{\omega}, \mathbf{b}) \in SE(3)$ is the Lumped Error \cite{richter2021robotic}, parameterized by an axis-angle vector $\boldsymbol{\omega} \in \mathbb{R}^3$ and a translational vector $\mathbf{b} \in \mathbb{R}^3$. 
The Lumped Error compensates for both errors in joint angles and hand-eye in real-time to precisely track the tool in the camera frame.
For more details, refer to our previous work \cite{richter2021robotic}.
Since the Lumped Error is not constant and the application is real-time tracking (i.e only detections up to time $t$ is known), a tracking formulation with a Hidden Markov Model is proposed where the posterior probability conditioned on all observations is estimated recursively:
}
\begin{multline}
\textcolor{black}{
    P( \boldsymbol{\omega}_t, \mathbf{b}_t | \mathbf{h}_{1:K, 1:t}) \propto  }\\ 
    \textcolor{black}{ 
    P( \mathbf{h}_{1:K, t} | \boldsymbol{\omega}_t, \mathbf{b}_t ) \int P(\boldsymbol{\omega}_t, \mathbf{b}_t | \boldsymbol{\omega}_{t-1}, \mathbf{b}_{t-1}) }\\ 
    \textcolor{black}{ 
    P( \boldsymbol{\omega}_{t-1}, \mathbf{b}_{t-1} | \mathbf{h}_{1:K, 1:t-1}) d [ \boldsymbol{\omega}_{t-1}, \mathbf{b}_{t-1}]^T}
\end{multline}
\textcolor{black}{
which is solved using a particle filter \cite{richter2021robotic}.
The motion and observation models are respectively defined as:}
\begin{equation}
\textcolor{black}{
    [\boldsymbol{\omega}_{t}, \mathbf{b}_{t}]^T \sim \mathcal{N}([\boldsymbol{\omega}_{t-1}, \mathbf{b}_{t-1}]^T, \mathbf{\Sigma_{\boldsymbol{\omega},b}} )
}
\end{equation}
\textcolor{black}{
where $\mathbf{\Sigma_{\boldsymbol{\omega},b}}$ is the covariance matrix and}
\begin{equation}
\textcolor{black}{
    P( \mathbf{h}_{{1:K},t}| \boldsymbol{\omega}_{t}, \mathbf{b}_{t})  \propto \sum \limits_{i=1}^K \rho_{i,t} e^{-\alpha ||\mathbf{h}_{i,t} - \mathbf{m}_i(\boldsymbol{\omega}_{t}, \mathbf{b}_{t}) ||^2}
    }
    \label{eq:observation_model_equation}
\end{equation}
\textcolor{black}{
where $\rho_{i,t}$ is the confidence score from the keypoint detector, and $\alpha$ is a tuning parameter.}

\begin{figure}[t]
\vspace{2mm}
\begin{subfigure}{0.24\textwidth}
\includegraphics[width=1\textwidth]{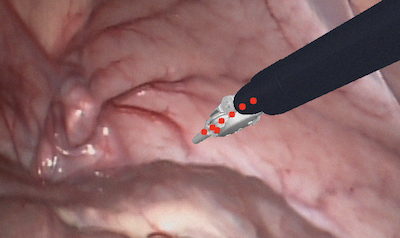}
\vspace{-0.12in}
\end{subfigure}
\begin{subfigure}{0.24\textwidth}
\includegraphics[width=1\textwidth]{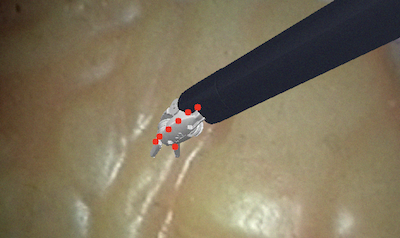}
\vspace{-0.12in}
\end{subfigure}
\\
\begin{subfigure}{0.24\textwidth}
\includegraphics[width=1\textwidth]{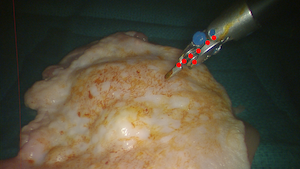}
\vspace{-0.12in}
\end{subfigure}
\begin{subfigure}{0.24\textwidth}
\includegraphics[width=1\textwidth]{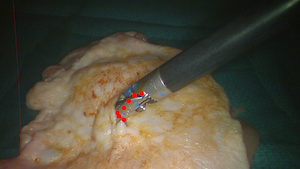}
\vspace{-0.12in}
\end{subfigure}
\caption{Detection of the optimized keypoints for the surgical tool on synthetic (top) and real (bottom) images. The 3D position of the optimized keypoints relative to the tool is shown in the middle-right of Fig. \ref{fig:keypoints_visualization}, and the DNN accurately predicts their projections on the image plane.}
\label{fig:super_keypoint_prediction}
\vspace{-0.12in}
\end{figure}

Differing from \cite{li2020super} and \cite{lu2020super}, we are using the optimized keypoints (red points in Fig. \ref{fig:keypoints_visualization}) instead of the hand-picked keypoints. Then, domain randomization technique is used to bridge the reality gap between the synthetic and real-world images. The resulting keypoint detections are shown in Fig. \ref{fig:super_keypoint_prediction}. Note that the optimized keypoints are inside of the tool body, and the DNN can accurately predict their projections onto the image plane in different tool configurations.

\begin{figure}[t]
\vspace{2mm}
    \centering
    \includegraphics[width=1.0\linewidth]{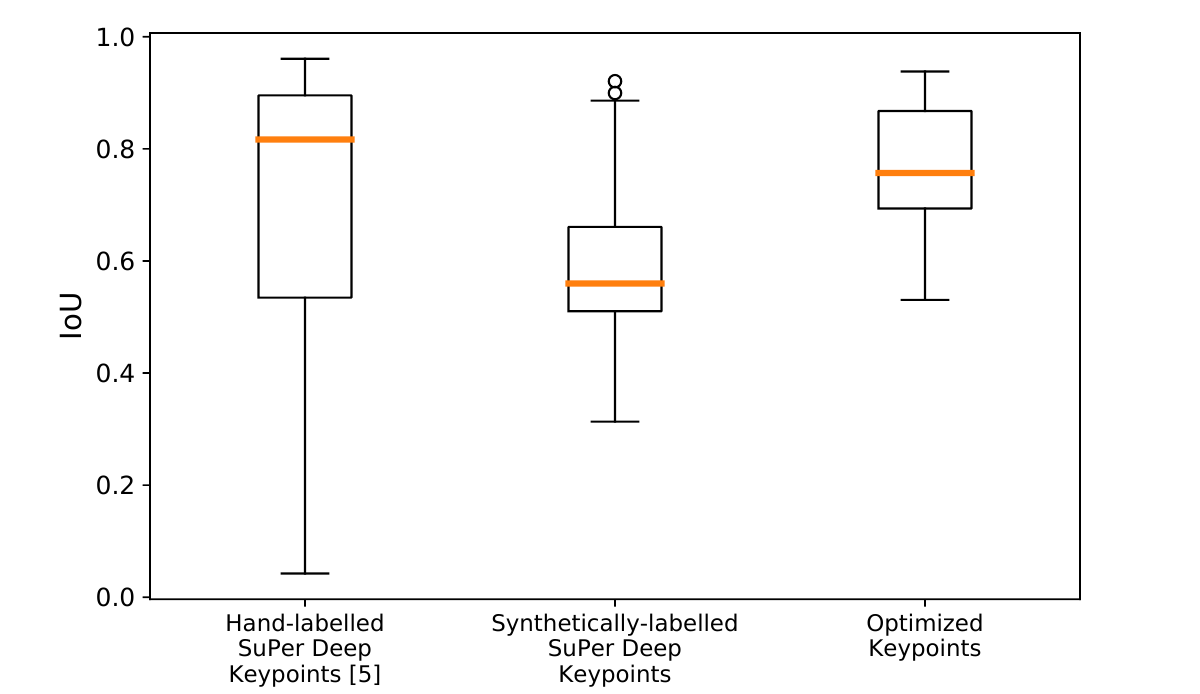}
    \caption{The box plot of the IoUs of the rendered tool mask using three different tool tracking setups (circles are outliers). The \textit{Optimized Keypoints} has less variance with high accuracy.}
    \label{fig:IoUs}
    \vspace{-0.12in}
\end{figure}

\begin{figure}[t]
\vspace{2mm}
\begin{subfigure}{0.24\textwidth}
\includegraphics[width=1\textwidth]{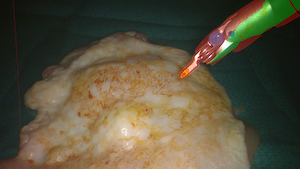}
\vspace{-0.12in}
\end{subfigure}
\begin{subfigure}{0.24\textwidth}
\includegraphics[width=1\textwidth]{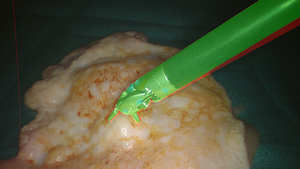}
\vspace{-0.12in}
\end{subfigure}
\\
\begin{subfigure}{0.24\textwidth}
\includegraphics[width=1\textwidth]{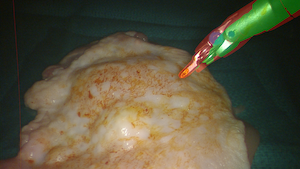}
\vspace{-0.12in}
\end{subfigure}
\begin{subfigure}{0.24\textwidth}
\includegraphics[width=1\textwidth]{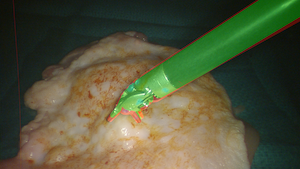}
\vspace{-0.12in}
\end{subfigure}
\\
\begin{subfigure}{0.24\textwidth}
\includegraphics[width=1\textwidth]{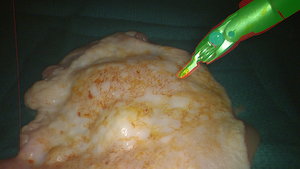}
\vspace{-0.12in}
\end{subfigure}
\begin{subfigure}{0.24\textwidth}
\includegraphics[width=1\textwidth]{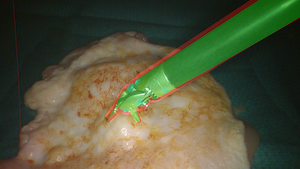}
\vspace{-0.12in}
\end{subfigure}
\caption{Qualitative results of the tool tracking for three different setups. 
From top to bottom, each row shows the results of \textit{Hand-labelled SuPer Deep Keypoints}, the \textit{Synthetically-labelled SuPer Deep Keypoints}, and the \textit{Optimized Keypoints}. The green area shows the intersection of the rendered mask and ground-truth mask ($\mathbf{G} \cap \mathbf{P}$), and the red area shows the difference between the rendered mask and ground-truth mask ($\mathbf{G} \cup \mathbf{P} - \mathbf{G} \cap \mathbf{P}$).}
\label{fig:tool_tracking_mask}
\vspace{-0.12in}
\end{figure}

For comparison, we evaluated the tool tracking performance with three different setups described as follows. 

\textit{Hand-labelled SuPer Deep Keypoints}: 
This setup is identical to the tool tracking approach in \cite{lu2020super}. The seven hand-picked keypoints were used for tracking, which is on the surface of the tool. The DNN was then trained on the 100 real-world images with the hand-labeled ground-truth position. 

\textit{Synthetically-labelled SuPer Deep Keypoints}: 
The second setup used the same set of hand-picked keypoints from SuPer Deep, but the DNN was trained on the around 20K synthetic images with domain randomization, where the keypoint labels are provided even with occlusions.

\textit{Optimized Keypoints}: The third setup was using the seven optimized keypoints, as shown in Fig. \ref{fig:keypoints_visualization}, and the DNN was also trained on the synthetic images with domain randomization. 

The tool tracking performance is computed by rendering a re-projected tool mask on the image frame from the estimation based on the keypoints, and the IoU is computed with the ground-truth mask for evaluation.
Quantitative and qualitative results are shown in Fig. \ref{fig:IoUs} and Fig. \ref{fig:tool_tracking_mask} respectively.
The setup with \textit{Hand-labelled SuPer Deep Keypoints} fails to track the tool when the tool was turning, as those hand-picked keypoints on the tool surface are occluded and humans cannot provide the label for those non-visible keypoints.
However, using the optimized keypoints, the DNN makes accurate predictions for non-visible keypoints, as shown in Fig. \ref{fig:super_keypoint_prediction}, since the synthetically generated training data can provide labels for those scenarios.
Although both \textit{Synthetically-labelled SuPer Deep Keypoints} and \textit{optimized keypoints} setups are utilizing the synthetic dataset, the \textit{optimized keypoints} achieves higher accuracy because the keypoints are optimizing the detection performance of the DNN.
Another advantage of the optimized keypoints is to reduce the ambiguity in detection.
As stated in \cite{lu2020super}, the detection of some keypoints is challenging due to the tool's symmetry, which causes false detections.
The optimized keypoints are instead distributed in an asymmetric pattern as shown in Fig. \ref{fig:keypoints_visualization}.

\section{Discussion and Conclusion}

We proposed a general keypoint optimization algorithm to maximize the performance of 2D and 3D localization on robotic manipulators. Our algorithm utilized a DNN for keypoint detections and can even handle self-occlusions by optimizing the keypoint locations and training on synthetically generated data.
\textcolor{black}{
The results show that the optimized keypoints yield higher accuracy on detection and localization compared to manually or randomly selected keypoints, hence resulting in better performance for the wide breadth of robotic applications that rely on keypoints for visual feedback.}
To show this, we presented both quantitative and qualitative results from two robotic applications: camera-to-base pose estimation and surgical tool tracking.
\textcolor{black}{
The experimental results of detecting optimized keypoints in cases of self-occlusion further motivate the importance of this work as previously manually selected keypoints were unable to produce this type of result.
For future work, we will incorporate the optimized keypoints for visual servoing and explore optimizing keypoints for other robot applications (e.g. motion planing \cite{das2020learning}).}

\balance
\bibliographystyle{ieeetr}
\bibliography{references}

\begin{thebibliography}{10}

\bibitem{fassi2005hand}
I.~Fassi and G.~Legnani, ``Hand to sensor calibration: A geometrical
  interpretation of the matrix equation ax= xb,'' {\em Journal of Robotic
  Systems}, vol.~22, no.~9, pp.~497--506, 2005.

\bibitem{qian2002online}
J.~Qian and J.~Su, ``Online estimation of image jacobian matrix by kalman-bucy
  filter for uncalibrated stereo vision feedback,'' in {\em Proceedings 2002
  IEEE International Conference on Robotics and Automation}, vol.~1,
  pp.~562--567, IEEE, 2002.

\bibitem{garrido2014aruco}
S.~Garrido-Jurado {\em et~al.}, ``Automatic generation and detection of highly
  reliable fiducial markers under occlusion,'' {\em Pattern Recognition},
  vol.~47, no.~6, pp.~2280--2292, 2014.

\bibitem{olson2011apriltag}
E.~Olson, ``Apriltag: A robust and flexible visual fiducial system,'' in {\em
  2011 IEEE International Conference on Robotics and Automation},
  pp.~3400--3407, IEEE, 2011.

\bibitem{lu2020super}
J.~Lu {\em et~al.}, ``Super deep: A surgical perception framework for robotic
  tissue manipulation using deep learning for feature extraction,'' {\em arXiv
  preprint arXiv:2003.03472}, 2020.

\bibitem{lee2020dream}
T.~E. Lee {\em et~al.}, ``Camera-to-robot pose estimation from a single
  image,'' in {\em 2020 IEEE International Conference on Robotics and
  Automation (ICRA)}, pp.~9426--9432, IEEE, 2020.

\bibitem{robotics2013baxter}
R.~Robotics, ``Baxter,'' {\em Retrieved Jan}, vol.~10, p.~2014, 2013.

\bibitem{dvrk}
P.~{Kazanzides} {\em et~al.}, ``An open-source research kit for the da vinci®
  surgical system,'' in {\em 2014 IEEE International Conference on Robotics and
  Automation (ICRA)}, pp.~6434--6439, May 2014.

\bibitem{lowe2004distinctive}
D.~G. Lowe, ``Distinctive image features from scale-invariant keypoints,'' {\em
  International journal of computer vision}, vol.~60, no.~2, pp.~91--110, 2004.

\bibitem{bay2006surf}
H.~Bay {\em et~al.}, ``Surf: Speeded up robust features,'' in {\em European
  conference on computer vision}, pp.~404--417, Springer, 2006.

\bibitem{nannen2013grid}
V.~Nannen and G.~Oliver, ``Grid-based spatial keypoint selection for real time
  visual odometry.,'' in {\em ICPRAM}, pp.~586--589, 2013.

\bibitem{tamimi2006localization}
H.~Tamimi {\em et~al.}, ``Localization of mobile robots with omnidirectional
  vision using particle filter and iterative sift,'' {\em Robotics and
  Autonomous Systems}, vol.~54, no.~9, pp.~758--765, 2006.

\bibitem{buoncompagni2015saliency}
S.~Buoncompagni {\em et~al.}, ``Saliency-based keypoint selection for fast
  object detection and matching,'' {\em Pattern Recognition Letters}, vol.~62,
  pp.~32--40, 2015.

\bibitem{mukherjee2016salient}
P.~Mukherjee, S.~Srivastava, and B.~Lall, ``Salient keypoint selection for
  object representation,'' in {\em 2016 Twenty Second National Conference on
  Communication (NCC)}, pp.~1--6, IEEE, 2016.

\bibitem{thewlis2017unsupervised}
J.~Thewlis {\em et~al.}, ``Unsupervised learning of object landmarks by
  factorized spatial embeddings,'' in {\em Proceedings of the IEEE
  international conference on computer vision}, pp.~5916--5925, 2017.

\bibitem{jakab2018unsupervised}
T.~Jakab {\em et~al.}, ``Unsupervised learning of object landmarks through
  conditional image generation,'' in {\em Advances in neural information
  processing systems}, pp.~4016--4027, 2018.

\bibitem{zhang2018unsupervised}
Y.~Zhang {\em et~al.}, ``Unsupervised discovery of object landmarks as
  structural representations,'' in {\em Proceedings of the IEEE Conference on
  Computer Vision and Pattern Recognition}, pp.~2694--2703, 2018.

\bibitem{suwajanakorn2018discovery}
S.~Suwajanakorn {\em et~al.}, ``Discovery of latent 3d keypoints via end-to-end
  geometric reasoning,'' in {\em Advances in neural information processing
  systems}, pp.~2059--2070, 2018.

\bibitem{schmidt2014dart}
T.~Schmidt, R.~A. Newcombe, and D.~Fox, ``Dart: Dense articulated real-time
  tracking.,'' in {\em Robotics: Science and Systems}, vol.~2, Berkeley, CA,
  2014.

\bibitem{cifuentes2016probabilistic}
C.~G. Cifuentes {\em et~al.}, ``Probabilistic articulated real-time tracking
  for robot manipulation,'' {\em IEEE Robotics and Automation Letters}, vol.~2,
  no.~2, pp.~577--584, 2016.

\bibitem{richter2021robotic}
F.~Richter, J.~Lu, R.~K. Orosco, and M.~C. Yip, ``Robotic tool tracking under
  partially visible kinematic chain: A unified approach,'' {\em arXiv preprint
  arXiv:2102.06235}, 2021.

\bibitem{lambrecht2019towards}
J.~Lambrecht and L.~K{\"a}stner, ``Towards the usage of synthetic data for
  marker-less pose estimation of articulated robots in rgb images,'' in {\em
  2019 19th International Conference on Advanced Robotics (ICAR)},
  pp.~240--247, IEEE, 2019.

\bibitem{mathis2018deeplabcut}
A.~Mathis {\em et~al.}, ``Deeplabcut: markerless pose estimation of
  user-defined body parts with deep learning,'' {\em Nature neuroscience},
  vol.~21, no.~9, p.~1281, 2018.

\bibitem{lepetit2009epnp}
V.~Lepetit {\em et~al.}, ``Epnp: An accurate o (n) solution to the pnp
  problem,'' {\em International journal of computer vision}, vol.~81, no.~2,
  p.~155, 2009.

\bibitem{coppeliaSim}
E.~Rohmer, S.~P.~N. Singh, and M.~Freese, ``Coppeliasim (formerly v-rep): a
  versatile and scalable robot simulation framework,'' in {\em Proc. of The
  International Conference on Intelligent Robots and Systems (IROS)}, 2013.
\newblock www.coppeliarobotics.com.

\bibitem{james2019pyrep}
S.~James {\em et~al.}, ``Pyrep: Bringing v-rep to deep robot learning,'' {\em
  arXiv preprint arXiv:1906.11176}, 2019.

\bibitem{tobin2017domain}
J.~Tobin {\em et~al.}, ``Domain randomization for transferring deep neural
  networks from simulation to the real world,'' in {\em 2017 IEEE/RSJ
  International Conference on Intelligent Robots and Systems (IROS)},
  pp.~23--30, IEEE, 2017.

\bibitem{quattoni2009indoor}
A.~Quattoni and A.~Torralba, ``Recognizing indoor scenes,'' in {\em 2009 IEEE
  Conference on Computer Vision and Pattern Recognition}, pp.~413--420, IEEE,
  2009.

\bibitem{mountney2010three}
P.~Mountney, D.~Stoyanov, and G.-Z. Yang, ``Three-dimensional tissue
  deformation recovery and tracking,'' {\em IEEE Signal Processing Magazine},
  vol.~27, no.~4, pp.~14--24, 2010.

\bibitem{imgaug}
A.~B. Jung {\em et~al.}, ``{imgaug},'' 2020.
\newblock Online; accessed 01-Feb-2020.

\bibitem{yang2012pck}
Y.~Yang and D.~Ramanan, ``Articulated human detection with flexible mixtures of
  parts,'' {\em IEEE transactions on pattern analysis and machine
  intelligence}, vol.~35, no.~12, pp.~2878--2890, 2012.

\bibitem{mobilenet}
M.~Sandler {\em et~al.}, ``Mobilenetv2: Inverted residuals and linear
  bottlenecks,'' {\em 2018 IEEE/CVF Conference on Computer Vision and Pattern
  Recognition}, Jun 2018.

\bibitem{richter2019sarpd}
F.~Richter {\em et~al.}, ``Augmented reality predictive displays to help
  mitigate the effects of delayed telesurgery,'' in {\em 2019 International
  Conference on Robotics and Automation (ICRA)}, pp.~444--450, IEEE, 2019.

\bibitem{richter2019dvrl}
F.~Richter {\em et~al.}, ``Open-sourced reinforcement learning environments for
  surgical robotics,'' {\em arXiv preprint arXiv:1903.02090}, 2019.

\bibitem{yipDasJournal}
M.~Yip and N.~Das, ``Robot autonomy for surgery,'' in {\em Encyclopedia of
  Medical Robotics}, ch.~10, pp.~281--313, World Scientific, 2017.

\bibitem{li2020super}
Y.~Li {\em et~al.}, ``Super: A surgical perception framework for endoscopic
  tissue manipulation with surgical robotics,'' {\em IEEE Robotics and
  Automation Letters}, vol.~5, no.~2, pp.~2294--2301, 2020.

\bibitem{das2020learning}
N.~Das and M.~Yip, ``Learning-based proxy collision detection for robot motion
  planning applications,'' {\em IEEE Transactions on Robotics}, vol.~36, no.~4,
  pp.~1096--1114, 2020.

\end{thebibliography}

\end{document}